# Calibration of the underlying surface parameters for urban flood using latent variables and adjoint equation

Running title: **Adjoint-Based Urban Flood Calibration**

Yongfu Tian, Shan Ding, Guofeng Su* and Jianguo Chen*

Affiliation: School of Safety Science, Tsinghua University, Beijing, 100084, China.

## Acknowledgments

Funding: This work was supported by the National Key Research and Development Program of China [grant numbers 2021YFC3001604]; and the National Natural Science Foundation of China [grant numbers 72293571].

# Calibration of the underlying surface parameters for urban flood using latent variables and adjoint equation

## Abstract

Calibrating the urban underlying surface parameters is crucial for urban flood simulation. We formulate the parameter calibration problem into an optimization problem within the Bayesian framework using the maximum likelihood principle. We adopt the urban flood dynamical system model as the surrogate model and innovatively introduce latent variables inspired by machine learning to represent more uncertainties, which can also be compatible with common physical parameter calibration. For more efficient optimization, we construct the adjoint equation of the surrogate model to obtain gradient information and propose the parameter sharing technique and the localization technique to reduce the computation complexity of the adjoint equation. A simple case verifies the proposed method can converge quickly and is insensitive to the observation time interval. In the case derived from Test 8A, we calibrate Manning's coefficient of urban roads, with a maximum relative error of 13.88% and a minimum of 1.16%.

# 1. Introduction

With the ongoing global climate change, the frequency and intensity of flood disasters are escalating globally (IPCC, 2021). Meanwhile, as urbanization progresses and industrial clusters develop, the losses resulting from urban floods are growing increasingly severe (ADREM, 2021). Hence, urban flood management assumes particular significance. One of the key technologies implicated is urban flood simulation, which can be employed for Low-Impact Development (Baek et al., 2015) and Stormwater Control Measures (Alamdari & Sample, 2019).

The urban underlying surface is highly complex, featuring artificial facilities such as roadways, buildings, and communities, etc., which exerts a significant influence on the hydrological process (Mei et al., 2024), and constitutes key input data for urban flood simulation technology. To consider the urban underlying surface in detail, there are two approaches: the first is to incorporate as much urban data as feasible when constructing models and setting boundary conditions, which is a prior approach; the second is to infer or calibrate urban underlying surface parameters based on the observed rainfall-runoff processes, which is a posterior approach. The combination of the prior and posterior approaches can establish a more appropriate urban flood prediction model.

Let's assume that the urban flood process can be fully described by a complete model and complete boundary conditions. Given that the complete model and the complete boundary conditions are not fully comprehended, surrogate models and surrogate boundary conditions are generally adopted in engineering practice. In order to absorb as much urban system data as possible and reduce the computational complexity, this paper adopts the Urban Flood Dynamical System model (UFDS) as the surrogate model. The UFDS model is a constructive one and pertains to a fully distributed hydrological model. On the one hand, this model can priorly absorb urban underlying surface data. On the other hand, its form constitutes a system of ordinary differential equations, facilitating the construction of the adjoint equation and thereby obtaining the gradient information of the model parameters. The detailed description of the model is presented in Section 2.2.

The calibration of the urban underlying surface parameters based on observed rainfall runoff data pertains to the realm of data assimilation. Generally, there are two data assimilation frameworks, namely the variational data assimilation framework and the Bayesian framework, and there are some intersections between the two (Law et al., 2015). The current parameter calibration studies predominantly rely on the variational data assimilation framework, within which the parameter calibration is converted into a single or multi-objective optimization problem and then addressed by diverse optimization approaches, such as pattern search, genetic algorithms, particle swarm algorithms, grid search, complex algorithm, and reinforcement learning (Jeung et al., 2023). For instance, Barco et al. optimized the parameters of the SWMM model through a complex algorithm, with the objective function being the weighted sum of the total flow volume, peak flow rate, and instantaneous flow rate at the outlet hydrograph (Barco et al., 2008). Madsen optimized the 9 parameters of the MIKE11/NAM hydrological model by using a stochastic complex algorithm, where the optimization objectives encompassed the overall water balance, overall shape of the hydrograph, peak flows, and low flows (Madsen, 2000).

Owing to the intricacy of the model, the parameter calibration problem within the variational framework is currently predominantly investigated through gradient-free optimization algorithms, which possess relatively low optimization efficiency. Moreover, gradient-free optimization algorithms frequently encounter the "curse of dimensionality" problem when addressing the high-dimensional parameter calibration problem. In parameter calibration studies, gradient-based optimization algorithms are often feared to get trapped in local minima (Behrouz et al., 2020; Mancipe-Munoz et al., 2014). Alongside the rapid advancement of machine learning, however, diverse optimization techniques have significantly mitigated the problem of gradient-based algorithms being prone to fall into local minima, such as the addition of momentum terms and the adjustment of the learning rate. Further, gradient-based algorithms are powerful tools for resolving high-dimensional optimization problems, such as the vast number of parameters in deep learning models. Additionally, the gradient information furnished by gradient-based algorithms directly reflects the sensitivity of the parameters. Consequently, gradient-based algorithms might possess an edge. For instance, Pujol et al. (Pujol et al., 2024) employed a differentiable two-dimensional shallow water hydraulic model to generate high-resolution sensitivity maps of local gradients and Derivative-Based Global Sensitivity Measures, thereby achieving the inference of high-dimensional multi-variable parameters using multi-source observational data. So, this paper will investigate the calibration effects of urban underlying surface parameters based on the gradient-based algorithms.

The latent variable approach is frequently employed in machine learning and deep learning, exemplified by the Hidden Markov model (Grewal et al., 2019) and the Variational Autoencoder (Kingma & Welling, 2013). Latent variables are typically used to capture the underlying structure or hidden patterns in data, thereby simplifying the learning process of the model and enhancing its expressive power. Relevant research on parameter calibration for urban flood models mainly focuses on calibrating physical quantities within the model, such as Manning's coefficient and infiltration decay rate. Nevertheless, numerous uncertainties exist in reality, which are attributed to the calibrated physical quantities, resulting in distorted physical quantities. Hence, this paper introduces the latent variable approach from the domain of machine learning to calibrate the state of latent variables, utilizing latent variables to absorb various uncertainties. Moreover, this approach is compatible with the calibration of common physical quantities.

In summary, within the parameter calibration of urban flood underlying surface, we adopt the Urban Flood Dynamical System model as the surrogate model and incorporate latent variables into the model. Subsequently, we formulate the adjoint equation of the UFDS model to obtain the gradient information and calibrate the latent variables by employing gradient-based optimization algorithms.

# 2. Method

### *2.1 Parameter calibration framework*

The parameter calibration problem can be regarded as a subproblem of data assimilation, and we will introduce our parameter calibration method within the Bayesian framework, with the principal idea coming from the work of Law, Stuart et al. (Law et al., 2015).

The urban flood surrogate model is denoted as $\mathcal{M}(h,t|h^0,z)$, where $h$ is the vector of water depths, $z$ is the latent random variables, $t$ indicates time, and $h^0$ is the given initial water depth vector. The surrogate model is abbreviated as $h(t)=\mathcal{M}(t,z)$. The observation operator is denoted as $O$, and the discrete time sequence of observation is $0<t_1<t_2<\ldots<t_K=T$, with the observation value random variables sequence denoted as $y=[y_1,\ldots,y_k,\ldots,y_K]$, where $y_k$ is the observed value vector at time $k$. Within the Bayesian framework, it holds that:

$$\mathbb{P}(z|y)=\frac{\mathbb{P}(y|z)\mathbb{P}(z)}{\mathbb{P}(y)}\propto\mathbb{P}(y|z)\mathbb{P}(z) \tag{1}$$

where $\mathbb{P}(y|z)$ represents the likelihood distribution of $y$ given $z$, $\mathbb{P}(z)$ denotes the prior distribution of $z$, and $\mathbb{P}(z|y)$ indicates the posterior distribution of $z$ upon receiving information of $y$. Suppose that the surrogate model $\mathcal{M}(h,t|h^0,z)$ satisfies the Markov property in the time direction. The prediction value sequence of the surrogate model at the observation locations is denoted as

$$\hat{y}=[\hat{y}_1,\hat{y}_2,\ldots,\hat{y}_K]=\left[O\big(\mathcal{M}(t_1,z)\big),O\big(\mathcal{M}(t_2,z)\big),\ldots,O\big(\mathcal{M}(t_K,z)\big)\right] \tag{2}$$

Observations are inevitably accompanied by errors, and it is assumed that the observation errors follow a multivariate Gaussian distribution. Specifically, for any $k$, the random vector $y_k-\hat{y}_k$ is assumed to follow the multivariate Gaussian distribution with a mean of 0 and a covariance matrix $\Gamma$, where $\Gamma$ is a non-negative diagonal matrix. Denote the Probability Density Function (PDF) of the posterior distribution $\mathbb{P}(z|y)$ by $\rho(z|y)$, and that of the likelihood distribution $\mathbb{P}(y|z)$ by $\rho(y|z)$. Under the Markov property of the surrogate model and the errors assumption, the PDF of likelihood $\rho(y|z)$ can be expressed as

$$\rho(y|z)\propto\ \exp\left(-\frac{1}{2}\sum_{k=1}^{K}\left|y_k-O\big(\mathcal{M}(t_k,z)\big)\right|_{\Gamma}^{2}\right)=\exp\big(-\Phi(z;y)\big) \tag{3}$$

Let the prior PDF of the latent random variables $z$ be $\exp\big(-\Psi(z)\big)/Z_c$, where $Z_c$ is the normalization constant. According to Bayes' formula (1), we have

$$\rho(z|y) \propto \exp(-\Phi(z;y)) \exp(-\Psi(z))$$
$$= \exp(-\mathrm{I}(z;y))$$

(4)

where $\mathrm{I}(z;y) = \Phi(z;y) + \Psi(z)$.

Based on the principle of Maximum Likelihood, the optimized objective function is the posterior probability density $\rho(z|y)$. Take the Maximum Likelihood estimate of the posterior PDF $\rho(z|y)$, denoted as $z^*$, that is

$$z^* = \mathrm{argmax}_{z\in R^{|z|}} \ \rho(z|y)$$

(5)

Since $\rho(z|y) \propto \exp(-\mathrm{I}(z;y))$, it follows that

$$z^* = \mathrm{argmin}_{z\in R^{|z|}} \ I(z;y)$$

(6)

where

$$I(z;y) = \frac{1}{2}\sum_{k=1}^{K} \left| y_k - O\big(\mathcal{M}(t_k, z)\big) \right|_{\Gamma}^{2} + \Psi(z)$$

(7)

The gradient or subgradient of the objective function $I(z;y)$ with respect to the latent variables $z$ is given by

$$\nabla_z I(z;y) = -\sum_{k=1}^{K} \big(\nabla_h O \nabla_z \mathcal{M}(t_k, z)\big)^T \Gamma^{-1} \Big(y_k - O\big(\mathcal{M}(t_k, z)\big)\Big) + \nabla_z \Psi(z)$$

(8)

*2.2 Urban flood dynamical system model*

We adopt the Urban Flood Dynamical System model (UFDS) to incorporate urban data a priori. Let $h = [h_1, h_2, \ldots h_i, \ldots, h_N]^{\top}$ denotes the vector of water depths, where $h_i$ represents the water depth (m) of the i-th cell and $N$ is the total number of cells. The UFDS model is presented as:

$$\begin{cases} \dfrac{dh_i}{dt} = -\dfrac{\sum_{j\in NB_i} a_{ij}(h_i, h_j)}{s_i} + p_i(t) - I_i(h_i, t) + B_i(h_i, t) \\ h(0) = h^0 \end{cases}$$

(9)

where, $s_i$, $p_i(t)$, $I_i(h_i, t)$ and $B_i(h_i, t)$ respectively signify the area $(m^2)$, rainfall intensity ($m/s$), infiltration intensity ($m/s$), and boundary inflow intensity (such as drainage) ($m/s$) of the i-th cell, $NB_i$ is the set of neighbouring cells of the i-th cell, $a_{ij}(h_i, h_j)$ represents the outflow flux ($m^3/s$) from cell $i$ to cell $j$ (or is referred to as the outflow rule), and $h^0$ is the initial water depth vector at time 0. For more information about this model, please refer to the supplementary materials.

The model is required to be adjusted posteriorly informed the observed historical rainfall-runoff process data. The common approach is to calibrate the model parameters, such as the ground roughness coefficient, infiltration coefficient, etc., thereby enabling the model parameters to absorb various uncertainties, including model bias, inflow and outflow boundary condition bias, data bias of the underlying surface, and physical parameter bias of the underlying surface, etc.

This paper presents a latent variable approach to assimilate the diverse uncertainties in the simulation of urban flood process by latent variables. Simultaneously, the latent variable approach is compatible with the calibration of model parameters. There exist several ways for introducing latent variables, and in this paper, they are introduced at the links between cells, which not only reflect the uncertainty of physical parameters of the underlying surface but also capture the uncertainty of the interactions among cells. The formula (9) is rewritten as:

$$\frac{dh_i}{dt} = -\sum_{j \in NB_i} z_{ij} a_{ij}(h_i, h_j)/s_i + p_i(t) - I_i(h_i, t) + B_i(h_i, t) \tag{10}$$

where $z_{ij}$ is the introduced latent variable between cell $i$ and cell $j$. $z_{ij}$ can reflect the uncertainty of the ground roughness, the uncertainty of the cell elevation, whether it is interrupted or not, etc. Let

$$Z = (z_{ij})_{N \times N} = \begin{cases} z_{ij}, & j \in NB_i \\ 1, & j \notin NB_i \end{cases} \tag{11}$$

be designated as the latent variable matrix. Each adjacent edge is assigned a latent variable, and the direction is immaterial, thus $Z$ is a symmetric matrix. If an edge is completely determined, then $z_{ij}$ is regarded as a constant. The non-constant latent variables in the matrix $Z$ are composed into a vector, denoted as $\boldsymbol{z}$.

*2.3 Adjoint Equation*

Let the solution of the latent variable form of the UFDS model (10) be $h(t, \boldsymbol{z})$, which serves as a function of the latent variables $\boldsymbol{z}$ and time $t$. We can establish the adjoint equation of the dynamical system (10). $\nabla_{\boldsymbol{z}} h_i(t, \boldsymbol{z})$ satisfies the following differential equation:

$$\frac{\partial \nabla_{\boldsymbol{z}} h_i}{\partial t} = \nabla_{\boldsymbol{z}} \frac{\partial h_i(t, \boldsymbol{z})}{\partial t} = \nabla_{\boldsymbol{z}} \left( \sum_{j \in NB_i} - z_{ij} a_{ij}/s_i + p_i(t) - I_i(h_i, t) + B_i(h_i, t) \right)$$
$$= \sum_{j \in NB_i} -a_{ij} \nabla_{\boldsymbol{z}} z_{ij}/s_i + \left( \frac{\partial B_i}{\partial h_i} - \frac{\partial I_i}{\partial h_i} - \sum_{j \in NB_i} z_{ij} \frac{\partial a_{ij}}{\partial h_i}/s_i \right) \nabla_{\boldsymbol{z}} h_i - \sum_{j \in NB_i} z_{ij} \frac{\partial a_{ij}}{\partial h_j}/s_i \nabla_{\boldsymbol{z}} h_j \tag{12}$$

At the initial moment of 0, the rainfall has just initiated, and a runoff must go through the generation process before it can occur. Thus, it is natural that $\nabla_{\boldsymbol{z}} h_i(0) = \boldsymbol{0}$. Accordingly, the gradient $\nabla_{\boldsymbol{z}} h_i$ satisfies the dynamical system (13).

$$\begin{cases} \dfrac{d\nabla_{\boldsymbol{z}} h_i}{dt} = \sum_{j \in NB_i} - a_{ij} \nabla_{\boldsymbol{z}} z_{ij}/s_i + \left( \dfrac{\partial B_i}{\partial h_i} - \dfrac{\partial I_i}{\partial h_i} - \sum_{j \in NB_i} z_{ij} \dfrac{\partial a_{ij}}{\partial h_i}/s_i \right) \nabla_{\boldsymbol{z}} h_i \\ \qquad\qquad - \sum_{j \in NB_i} z_{ij} \dfrac{\partial a_{ij}}{\partial h_j}/s_i \nabla_{\boldsymbol{z}} h_j \\ \nabla_{\boldsymbol{z}} h_i(0) = \boldsymbol{0} \end{cases} \tag{13}$$

The dynamical system (13) is called the adjoint equation of the dynamical system (10). The dynamical system (10) can be merged with its adjoint equation (10) to constitute a system of ordinary differential equations, which can be numerically solved in its entirety.

To represent the adjoint equation (13) in a vectorized form, several notations are introduced. $\nabla_z h$ has a dimension of $N \times |\boldsymbol{z}|$, and for each cell $i$, a vector $E_{i*}$ with dimension $1 \times |\boldsymbol{z}|$ is constructed. If $z_s \in \boldsymbol{z}$ is on the adjacent edge $(i, j)$ of cell $i$, where $j \in NB_i$, then $E_{is} = 1$; otherwise, $E_{is} = 0$. The vectors $E_{i*}$ constitute the following matrix:

$$E = (E_{is})_{N \times |\boldsymbol{z}|} = \begin{cases} 1, & z_s \in NB_i \\ 0, & else \end{cases}$$

(14)

Introduce the outflow rule matrix $A$ and the partial derivative matrix $A'$ as follows:

$$A = (a_{ij})_{N\times N} = \begin{cases} a_{ij}, & j \in NB_i \\ 0, & j \notin NB_i \end{cases}, \qquad A' = \begin{bmatrix} \frac{\partial a_{11}}{\partial h_1} & \cdots & \frac{\partial a_{1N}}{\partial h_N} \\ \vdots & \ddots & \vdots \\ \frac{\partial a_{N1}}{\partial h_1} & \cdots & \frac{\partial a_{N1}}{\partial h_N} \end{bmatrix}$$

(15)

Denote $s = [s_1, s_2, \dots, s_N]^{\top}$. Define the operator $\sigma(x)$ as the mapping of a vector $x$ into a diagonal matrix. In both the UFDS model and its adjoint equation, the primary challenge resides in the convection term, represented by matrices $A$ and $A'$. To simplify the analysis, this study temporarily omits consideration of the infiltration and inflow terms. Then, the adjoint equation (13) can be rewritten as follows:

$$\begin{cases} \frac{d\nabla_{\boldsymbol{z}} h}{dt} = -AE/s + [\sigma((Z \otimes A')^T 1_e) - Z \otimes A']\nabla_{\boldsymbol{z}} h/s \\ \nabla_{\boldsymbol{z}} h(0) = \mathbf{0} \end{cases}$$

(16)

*2.4 Properties of the adjoint equation*

This section is devoted to deriving two properties of the adjoint equation (16), namely, the conservation of the total gradient and the diffusivity. The two properties will be utilized to develop localization technique.

First, let us analyse the conservation of the total gradient. By multiplying both sides of equation (16) by $1_e^T$ on the left, it can be observed that

$$\frac{d 1_e^T \nabla_{\boldsymbol{z}} h}{dt} = -1_e^T AE/s + 1_e^T[\sigma((Z \otimes A')^T 1_e) - Z \otimes A']\nabla_{\boldsymbol{z}} h/s = \mathbf{0}$$

(17)

It can be perceived that for any latent variable, the summation of the gradients of all cells within the cell field with respect to the latent variable remains constant over time, that is, the total gradient is conserved. The total gradient of the adjoint equation (16) is zero at the initial moment, and thus it is zero at any subsequent time.

Subsequently, the diffusivity is analysed. The adjoint equation (16) manifests a special diffusion process. For the sake of analysis convenience, a one-dimensional scenario is considered, where the cell $i$ has merely two neighboring cells with indices

$i-1$ and $i+1$ on the left and right, respectively. Then, $d(\nabla_{\boldsymbol{z}} h_i)/dt$ is expressed as

$$
\begin{aligned}
s_i \frac{d(\nabla_{\boldsymbol{z}} h_i)}{dt} = {} & a_{i-1,i} \nabla_{\boldsymbol{z}} z_{i-1,i} - a_{i,i+1} \nabla_{\boldsymbol{z}} z_{i,i+1} \\
& + \left[ \left( z_{i,i+1} \frac{\partial a_{i+1,i}}{\partial h_{i+1}} \nabla_{\boldsymbol{z}} h_{i+1} - z_{i,i+1} \frac{\partial a_{i,i+1}}{\partial h_i} \nabla_{\boldsymbol{z}} h_i \right) \right. \\
& \left. - \left( z_{i-1,i} \frac{\partial a_{i,i-1}}{\partial h_i} \nabla_{\boldsymbol{z}} h_i - z_{i-1,i} \frac{\partial a_{i-1,i}}{\partial h_{i-1}} \nabla_{\boldsymbol{z}} h_{i-1} \right) \right]
\end{aligned}
\tag{18}
$$

The right-hand side of the square bracketed item in Equation (18) bears resemblance to the difference format of the Laplace operator, and the $a_{i-1,i} \nabla_{\boldsymbol{z}} z_{i-1,i} - a_{i,i+1} \nabla_{\boldsymbol{z}} z_{i,i+1}$ constitutes the gradient source term. Equation (18) bears similarity to the heat conduction equation featuring anisotropic and time-varying thermal conductivity. Nevertheless, it differs from heat diffusion. Given the total gradient conservation and the fact that the total gradient is perpetually zero, the cell field concurrently exhibits positive and negative gradients in diffusion. Whenever water flows through a boundary with a latent variable, both a positive and a negative gradient of the latent variable are concurrently generated, with the absolute values being identical. The positive gradient is engendered by the cell below the flow direction of the boundary, and the negative gradient is produced from the cell above the flow direction of the boundary. Subsequently, the positive and negative gradients diffuse to adjacent cells, and when they encounter each other in the same cell, they neutralize one another. The gradient of any cell with respect to the latent variable is equivalent to the sum of its positive and negative gradients. For a cell $k$ that is not adjacent to the latent variable $z_s$, the gradient diffusion equation for $\nabla_{z_s} h_k$ lacks a source term, and all the gradient information it receives stems from the gradient diffusion of adjacent cells.

*2.5 Parameter Sharing and Localization Techniques*

In the adjoint equation (16), the dimension of $\nabla_{\boldsymbol{z}} h$ is $N \times |\boldsymbol{z}|$, where $N$ is the total number of cells and is typically quite large, and $|\boldsymbol{z}|$ is the number of latent variables, which is also considerable. Consequently, the numerical complexity of the adjoint equation (16) might be unacceptable. Evidently, there exist two paths to reduce the dimension of the equation: reducing $|\boldsymbol{z}|$ or reducing $N$. Following the two paths, we propose the parameter sharing technique and the localization technique respectively below.

**Parameter sharing technique**. The urban underlying surface can be partitioned into blocks, and the physical parameters within a block are identical, such as a road block or a grassland block. Hence, we introduce the parameter sharing technique, sharing the same latent variable for a block. As depicted in the left figure of Fig.1, the light blue area constitutes the parameter sharing domain, and the orange dashed lines represent the shared latent variables, whose value are the same. The grey dotted lines on each side of the edge are the independent latent variables.

If the city is partitioned into ω domains, the parameter sharing technique reduces the dimension of $\nabla_z h$ from $N \times |\boldsymbol{z}|$ to $N \times \omega$, thereby significantly decreasing the number of equations and the computational complexity.

**Localization technique**. Owing to the diffusivity of the adjoint equation (16), the gradient is generated from the cells adjacent to the latent variable and continuously spreads to the upstream and downstream cells. The gradient information possessed by a single cell situated farther away from the latent variable is less in comparison with that held by a single cell located in proximity. Nevertheless, the total number of cells located at a greater distance is significantly larger than that of the nearby cells, consuming a larger amount of computation. Based on this phenomenon, we introduce a localization technique, that is, considering only the cells in the vicinity of the latent variable in the adjoint equation (16) and disregarding the cells located far away.

The central red dashed line in the right figure of Fig.1 represents a single latent variable, and the positive and negative gradient information from the two adjacent cells above and below the latent variable spreads to neighbouring cells, with cells closer to the latent variable holding higher absolute gradient values. The green boxed area represents the localization domain, and the gray cells outside the green boxed area are the outer boundary. The localization technique merely calculates the adjoint equation within the localization domain and its outer boundary.

Parameter sharing technique and localization technique respectively decrease the dimension of the adjoint equation (16) in terms of the number of latent variables and the number of cells. When employed in combination, they are capable of reducing the computational complexity of the adjoint equation (16) to an acceptable level, and even enabling fast parameter calibration.

# 3. Cases and Results

### *3.1 Case 1: Simple Case*

We initially establish a simple case to observe, deliberate, and validate the diverse properties and calibration effects of the parameter calibration method proposed. Case 1 simulates a square area consisting of a total of 30*30 cells, with a cell edge length of 10m as depicted in Fig.2. The cell field is divided into four regions, labelled 1, 2, 3, and 4. Among them, regions 1 and 4 have a ground elevation of 1m, while regions 2 and 3 have a ground elevation of 0.8m.

**Latent variables configuration**. The latent variables are allocated to the four adjacent sides of the four regions, as indicated by the red lines in Fig.2. The latent variables are numbered successively. The latent variables along the red vertical lines are numbered from 0 to 29 from top to bottom, and those along the red horizontal lines are numbered from 30 to 59 from left to right, totalling 60 latent variables, which are mutually independent. The true values of the latent variables numbered 0 to 14 and 45 to 59 are set to 1.5, and the true values of the latent variables numbered 15 to 44 are set to 0 (i.e., wall), as shown in Fig2. The prior values of all latent variables are set to 1.0, which underestimates the true value of 1.5 and overestimates the true value of 0.

**Observation schemes configuration**. Case 1 sets four water level observation points, situated at the cells indicated by colour blocks 1, 2, 3, and 4 in Fig.2. This case concludes 4 types of observation point combinations, namely (2,3), (1,4), (1,2,3), and (1,2,3,4); and 2 types of observation interval combinations, namely 60 and 120 seconds. Eight observation schemes are further combined, as presented in Tab.1. For instance, (2,3; 60s) implies that the water level values at cells 2 and 3 are observed every 60 seconds.

**Rainfall settings**. In Case 1, four types of rainfall are designed, one being uniform rainfall, and the remaining three being Chicago type rainfall (Chen et al., 2023), with peak coefficient ratios of 0.3, 0.5, and 0.7 respectively. The intensity curves of the four rainfalls are depicted as solid lines in Fig.3. The duration of each rainfall lasts for 10 minutes, and the cumulative rainfall amount is 40 mm.

**Simulation**. The initial water depth is **0**. The outflow rule matrix $A(h)$ employs the Manning formula in the form of a windward scheme, and the water head flow between two adjacent cells is given by

$$a_{ij} = \left\{\frac{1}{2}\left[\left(1 + sign(sl_{ij})\right) h_i - \left(1 - sign(sl_{ij})\right) h_j\right]\right\}^{5/3} \frac{|sl_{ij}|^{1/2}}{n_{ij} s} \tag{19}$$

Herein $sl_{ij} = (h_i - h_j + e_i - e_j)/\Delta x$, where $e_*$ represents the elevation of the cell, $\Delta x$ denotes the distance between the centers of two cells, $n_{ij} = (n_i + n_j)/2$, where $n_*$ is the Manning's coefficient, $sign(*)$ is the sign function, and $s$ is the cell area. The elevation of Case 1 is presented in Fig.2, and the Manning's coefficient is uniformly set to $0.02 s/m^{1/3}$. $\Delta x$ is 10 m, and s is 100 $m^2$. The UFDS model (10) constitutes an initial value problem for a system of ordinary differential equations, which is numerically solved using the RK45 algorithm (Dormand & Prince, 1980) from the SciPy 1.14.1 package (Virtanen et al., 2020).

**Parameter calibration algorithm setup**. In Case 1, the optimizer selected is the Adagrad algorithm from Pytorch v2.4.1 (Paszke et al., 2019), with the learning rate (or step size) set at 0.2 and the total number of optimization steps set to 40. The prior distribution of each latent variable is configured as a uniform distribution on $[0, +\infty)$, signifying that the prior knowledge of the latent variable is merely its non-negativity. The water level observation error is hypothesized to be an independent and identically distributed Gaussian distribution with a variance of 0.001 and a mean of 0; that is, the covariance matrix $\Gamma$ in Equations (7) and (8) is the unit matrix multiplied by 0.001. At each optimization step, the forward solution of the UFDS model (10) and its adjoint equation (16) is required, and they are combined and solved numerically using RK45 (Dormand & Prince, 1980) in a synchronous manner.

**Training to convergence**. Under the aforementioned settings, the uniform rainfall depicted in Fig.3, namely the "uniform_1" line, is chosen as the training rainfall. The descent processes of the loss functions (i.e. objective function $I(z; y)$ in formula (7)) for the parameter calibration of the 8 observation schemes are presented in Fig.4. Within 40 iterations, the loss functions for the 8 observation schemes all converge to a local extremum. With the exception of C1T2 and C1T6, whose observation points are both (1,3), the other schemes converge super-linearly to the extremum within 10 steps.

**Insensitive to the observation time interval**. By comparing the convergence processes of the loss functions under the two observation time intervals of 60s and 120s, it is found that the trends of the two are essentially the same, and the difference in the extremum results from the disparity in the number of observation samples. Evidently, the method of the adjoint equation proposed for parameter calibration is insensitive to the observation time interval. This characteristic is of particular significance for real-world applications, given that the sampling interval of water level observation is typically within the minute range, such as 1 minute and 5 minutes, and the observation samples may have issues like missing values and extreme values due to limitations in signal acquisition, transmission, and storage.

**Generalization**. The three Chicago type rainfalls depicted by the solid lines in Fig.3, are selected as the test rainfalls. The test results (average of the three) are presented in Fig.4 on the right. For the sake of simplicity, only the representative C1T1 and C1T2 schemes are analysed. To assess the test effect, the Root Mean Square Error (RMSE) is employed as follows:

$$RMSE = \frac{1}{KJ}\sqrt{\sum_{k=1}^{K}\sum_{j=1}^{J}\left(x_{k,j} - \hat{x}_{k,j}\right)^2} \tag{20}$$

where $K$ represents the dimension of the time axis, $J$ represents the dimension of the spatial axis, $x$ denotes the true value, and $\hat{x}$ represents the prediction value. Regarding the loss functions, the C1T2 scheme converges to a lower extremum for both training and test rainfalls. Nevertheless, the total RMSE of the C1T2 scheme in test rainfalls remains high and does not exhibit a significant reduction, as depicted in Fig.4 on the right. Fig.5 on the right showcases the posterior values of the latent variables of the C1T2 scheme at each iteration, where the values of the latent variables from 0 to 14 and 45 to 59 are close to 0, which are far from the true value of 1.5. This indicates that the C1T2 scheme experiences overfitting. The RMSE of the water level at the observation point and in the overall cell field of the C1T1 scheme in test rainfalls both rapidly converge to the vicinity of 0, which is consistent with the trend of the loss function. As shown in Fig.5 on the left, the latent variables of the C1T1 scheme converge to the vicinity of the true value. By comparing the two schemes, two conclusions can be drawn: the convergence of the loss function to the extremum is a necessary but not sufficient condition for the parameter calibration to the true value; the selection of observation points has a significant influence on the parameter calibration results, and choosing inappropriate points may lead to overfitting.

**The advantages of the latent variables**. Regarding the blocking phenomena between regions 1 and 3 and between 4 and 3, if the Manning's coefficient is adopted as the calibration variable, extremely large or even infinite Manning's coefficients would be required, which renders the optimization process unstable. Nevertheless, the latent variables introduced in this paper converge rapidly during the optimization process and are capable of reflecting the numerous blocking phenomena existing in the city, such as walls, thereby conferring superiority over common direct parameter calibration.

**The gradient diffusion phenomenon**. The gradients of observation point 3 with respect to the latent variables 7 and 37 are respectively presented in the left and right figures of Fig.6. The positions of the latent variables 7 and 37 are shown in Fig.2. It is noted that there is a significant difference in the scale of the gradient coordinate axes in the two figures of Fig.6, and the gradient value of the latent variable 37 is considerably larger than that of the latent variable 7. In terms of distance, the distance from the observation point 3 to the latent variable 37 is less than half of the distance to the latent variable 7. Hence, it can be verified that the hypothesis that the gradient information held by cells farther from the latent variable is smaller, thereby supporting the localization technique.

### *3.2 Case 2: Urban Case*

In Case 2, the parameter calibration method along with the corresponding parameter sharing technique and localization technique proposed in this paper are applied to a real urban area to validate their efficacy. Case 2 employs the benchmark case Test 8A offered by the UK Environment Agency (EA). This benchmark case was utilized in the "Benchmarking the latest generation of 2D hydraulic modelling packages" report published by the EA to verify the accuracy of hydraulic models (Neelz

& Pender, 2010). Note that the setting of Case 2 in this paper is different from that in the original Test 8A. Case 2 constitutes a 0.4km*0.96km urban area featuring a left-low-right-high tendency, as depicted in Fig.7. Case 2 partitions it into 201*483 cells with a side length of 2m, amounting to a total of 97,083 cells. To explore the influence of the number of observation points on the calibration outcomes, two sets of observation schemes are established: the first encompasses 18 observation points, situated at the circular red points in Fig.7; the second set consists of 9 observation points, located at the yellow triangular points in Fig.7. The basic principle for the selection of observation points is to have them as evenly distributed as possible across each road area.

Urban roadways typically lie lower than the adjacent ground and constitute one of the principal routes for surface runoff, exerting a significant role in urban flooding events (Maksimovic et al., 2009). In Case 2, the Manning formula is adopted as the surface runoff formula, where the most crucial parameter is Manning's coefficient, as indicated in the equation (19). Hence, the parameter calibration objective for Case 2 is Manning's coefficient of all roads within the cell field, as depicted in the grey area of Fig.7. The true value of Manning's coefficient for non-road areas within the cell field is set at $0.05s/m^{1/3}$, while the true value for road areas is set at $0.02s/m^{1/3}$. In this case only the Manning's coefficient for road areas is calibrated; Manning's coefficient for non-road areas is a fixed value. In accordance with the proposed latent variable calibration method, Case 2 does not directly calibrate Manning's coefficient for road cells but calibrates the latent variables at the links of road cells. The Manning's coefficient for road cells is adjusted to $0.05s/m^{1/3}$, and the corresponding true value of the latent variables is set to 2.5, which is equivalent to the Manning's coefficient for the road cells being $0.02s/m^{1/3}$. The prior values of the latent variables are set to 1.0 and 4.0 to account for the overestimation and underestimation of the latent variables, respectively.

As depicted in Fig.7, the quantity of road cells is considerable, leading to a high dimension of latent variables. Given the relatively small area of this region and the standardization of the hardened road surface, it is hypothesized that all road cells possess the identical Manning's coefficient. Thus, the parameter sharing technique is employed to reduce all latent variables to a single latent variable for calibration, significantly lowering the computational complexity. Additionally, to further decrease the dimension of the adjoint equation (16), Case 2 adopts the localization technique and sets the localization region as the union of square areas with a side length of 8 (in terms of cell numbers) centred on the latent variables.

The simulated rainfall is a short-duration and uniform one, as indicated by the red dotted line "uniform_2" in Fig.3. The UFDS model and the HEC-RAS software (Suriya & Mudgal, 2012) SWE-EM (stricter momentum) solver are utilized to simulate the flooding process under this rainfall. By using the UFDS model as the simulator, the objective is to explore the effect of parameter calibration when the surrogate model is an accurate one, thereby controlling model bias and investigating the influence of other modules on parameter calibration. The HEC-RAS model is widely employed in urban flood simulation projects; thus, in this case, it is utilized as an approximate generator of the real rainfall runoff process.

Similar to Case 1, the initial water depth is set to $\mathbf{0}$. The outflow rule matrix $A(h)$ within the UFDS model also employs the Manning formula, as presented in equation (19). The simulation duration is set to 300s, and the observation time points are 60s, 120s, 180s, and 240s.

In Case 2, the optimizer is configured as the RMSprop algorithm from the PyTorch library (Paszke et al., 2019), with a learning rate of 0.1 and the total number of optimization steps of 40. The latent variable possesses no prior information. The error covariance matrix for the observations is assigned to the identity matrix multiplied by 0.001. The observation operator is defined as the single cell observation. The UFDS model (10) and its adjoint equation (16) are still numerically solved in a coupled manner using the RK45 algorithm (Dormand & Prince, 1980).

Two sets of prior values for the latent variable (1.0 and 4.0), two simulators (DFDS and HEC-RAS), and two sets of observation points (18 and 9) jointly combine to constitute 8 schemes, respectively designated as UFDS_1_18, UFDS_4_18, UFDS_1_9, UFDS_4_9, HEC-RAS_1_18, HEC-RAS_4_18, HEC-RAS_1_9, and HEC-RAS_4_9. The loss functions for the training process of the 8 schemes are depicted in the left figure of Fig.8, all of which have decreased to stable values. The lines fewer than 40 Epochs are because they have already converged, with the convergence criterion being a relative error of loss value less than $10^{-5}$. The loss values of the UFDS simulator have dropped to the vicinity of 0, while the loss values of the HEC-RAS simulator are higher due to the bias existing between the two models.

The posterior values of the latent variable in the road cells under the 8 schemes are presented in the right figure of Fig.8, where the posterior values of the latent variable converge approximately to 2.5 for all schemes. For those schemes that fail to satisfy the convergence conditions, the average value from the 30th to the 40th stage is adopted as the estimated value of the latent variable. The posterior values of the latent variable and their relative errors for the 8 schemes are presented in Tab.2. The reason why the posterior values of the latent variable in the HEC-RAS model scheme are greater than the true values might lie in the fact that the SWE-EM solver of the HEC-RAS software encompasses complete momentum terms, while the UFDS model attenuates the momentum, thereby increasing the latent variable value to compensate for the attenuation of momentum.

**Efficacy analysis**. It can be inferred from Tab.2 that if the UFDS model serves as both a simulator and a calibration model, the parameter calibration outcomes are nearly in accordance with the actual values, which can validate the efficacy of the parameter sharing technique, the localization technique, and the overall parameter calibration algorithm proposed in this paper. With HEC-RAS utilized as the actual rainfall runoff generator and the UFDS model along with its adjoint equation as the calibrator, the parameter calibration results are relatively close to the actual values, with the maximum relative error being 13.88% and the minimum 1.16%. It is manifest that the approach of the adjoint equation of the surrogate model, the latent variable approach, the parameter sharing technique, and the localization technique proposed in this paper can effectively calibrate the urban underlying surface parameters.

**Computational complexity analysis**. If the parameter sharing technique and the localization technique are not employed, the original dimension of the adjoint equation (16) in Case 2 would be 97083*29138, which is unacceptable. By applying the parameter sharing technique, it is assumed that all road cells have the same Manning's coefficient, and the dimension of the adjoint equation is reduced to 97083*1. Further by using the localization technique, if the localization region is set as a rectangle of length 8, the dimension of the adjoint equation is decreased to 44048*1; if the localization region is solely the road area, the dimension of the adjoint equation is reduced to 14569*1. This indicates that the parameter sharing technique and the localization technique can significantly lower the dimension of the adjoint equation and the computational complexity.

# 4. Conclusion

This paper presents a parameter calibration method for urban underlying surface parameters based on the latent variable approach and the adjoint equation of the surrogate model, and validates it in a simple case and a city case, yielding the following conclusions:

By introducing the latent variable approach from machine learning into the calibration of urban underlying surface parameters, the results demonstrate that the latent variable method is not only compatible with common direct parameter calibration approach for physical quantities but also capable of absorbing various uncertainties, such as topographic barriers and excessive flows.

The adjoint equation of the UFDS model is constructed, and the conservation of the total gradient and diffusivity of the adjoint equation are derived. Based on these two attributes, the parameter sharing technique and the localization technique were proposed, significantly reducing the dimension and the computational complexity of the adjoint equation.

The parameter calibration of urban underlying surface not only requires considering the training error but also the generalization error. The selection of observation locations has a substantial impact on the generalization effect. It cannot be regarded that a lower training error obtained in a certain observation scheme implies a superior calibration outcome.

Tab.1. The 8 observation schemes in Case 1

| Observation schemes | C1T1 | C1T2 | C1T3 | C1T4 | C1T5 | C1T6 | C1T7 | C1T8 |
|---|---|---|---|---|---|---|---|---|
| Observation points | 2,3 | 1,2 | 1,2,3 | 1,2,3,4 | 2,3 | 1,2 | 1,2,3 | 1,2,3,4 |
| Observation interval | 60s | 60s | 60s | 60s | 120s | 120s | 120s | 120s |

Tab.2. Posterior values of the latent variable and relative errors for the 8 schemes.

| Schemes | UFDS | | | | HEC-RAS | | | |
|---|---|---|---|---|---|---|---|---|
| | 1_18 | 4_18 | 1_9 | 4_9 | 1_18 | 4_18 | 1_9 | 4_9 |
| Posterior values | 2.499 | 2.618 | 2.499 | 2.5 | 2.833 | 2.847 | 2.834 | 2.529 |
| Relative errors | 0.04% | 4.72% | 0.04% | 0 | 13.32% | 13.88% | 13.36% | 1.16% |
| Manning's coefficient | 0.02 | 0.019 | 0.02 | 0.02 | 0.018 | 0.018 | 0.018 | 0.02 |

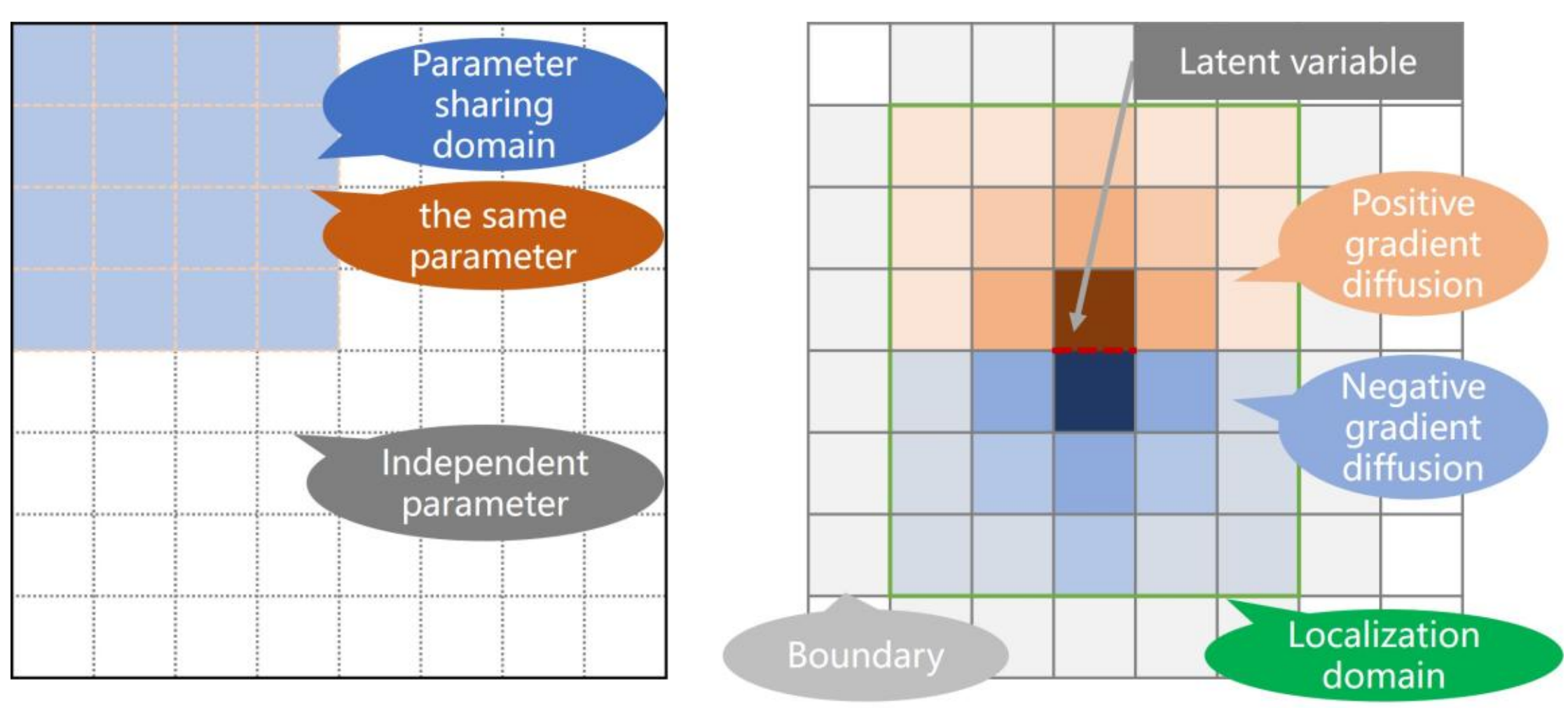

Fig.1. Illustration of parameter sharing technique (left) and localization technique (right)

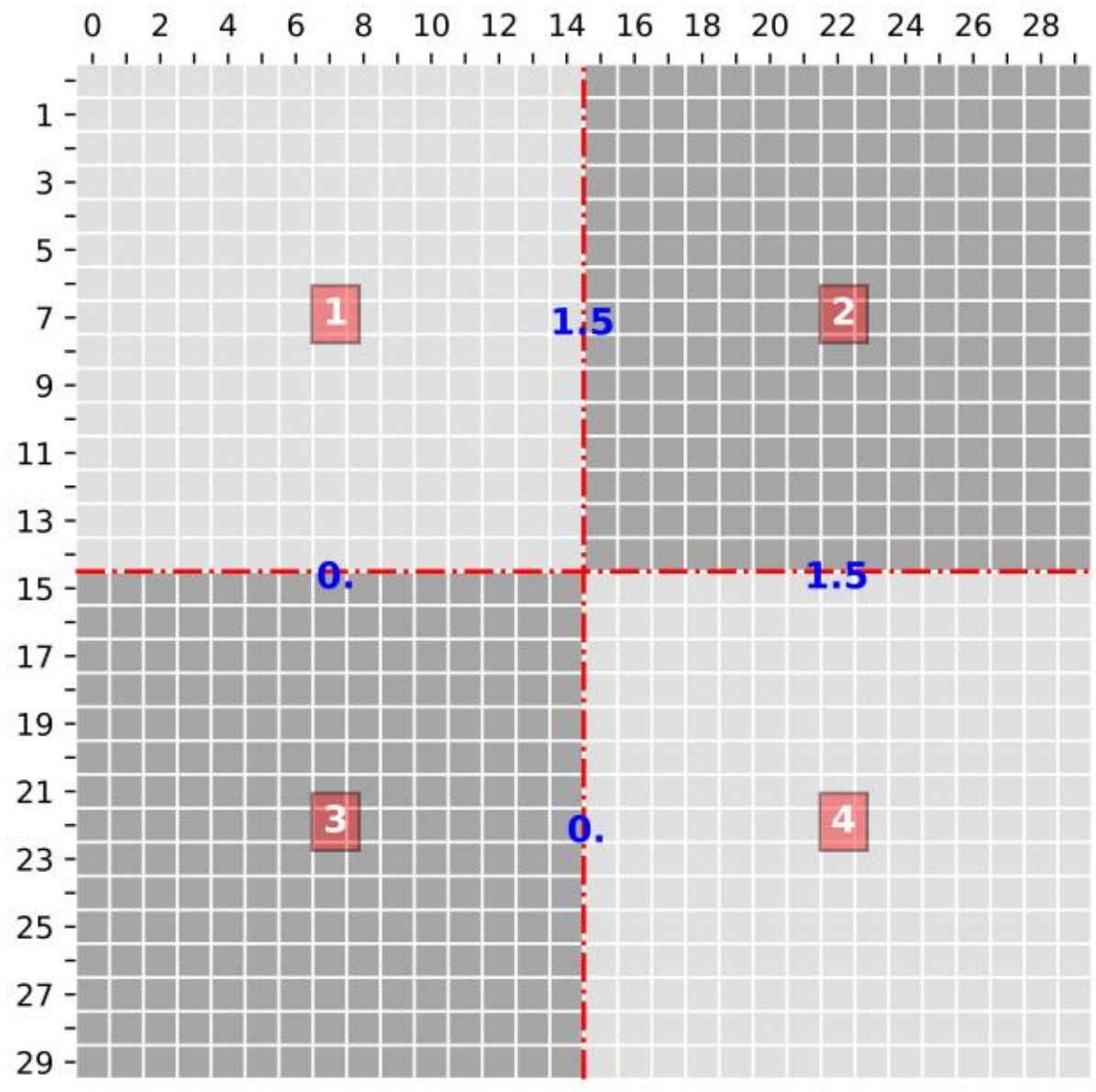

Fig.2. Configuration of simulation area, latent variables, and observation points for Case 1

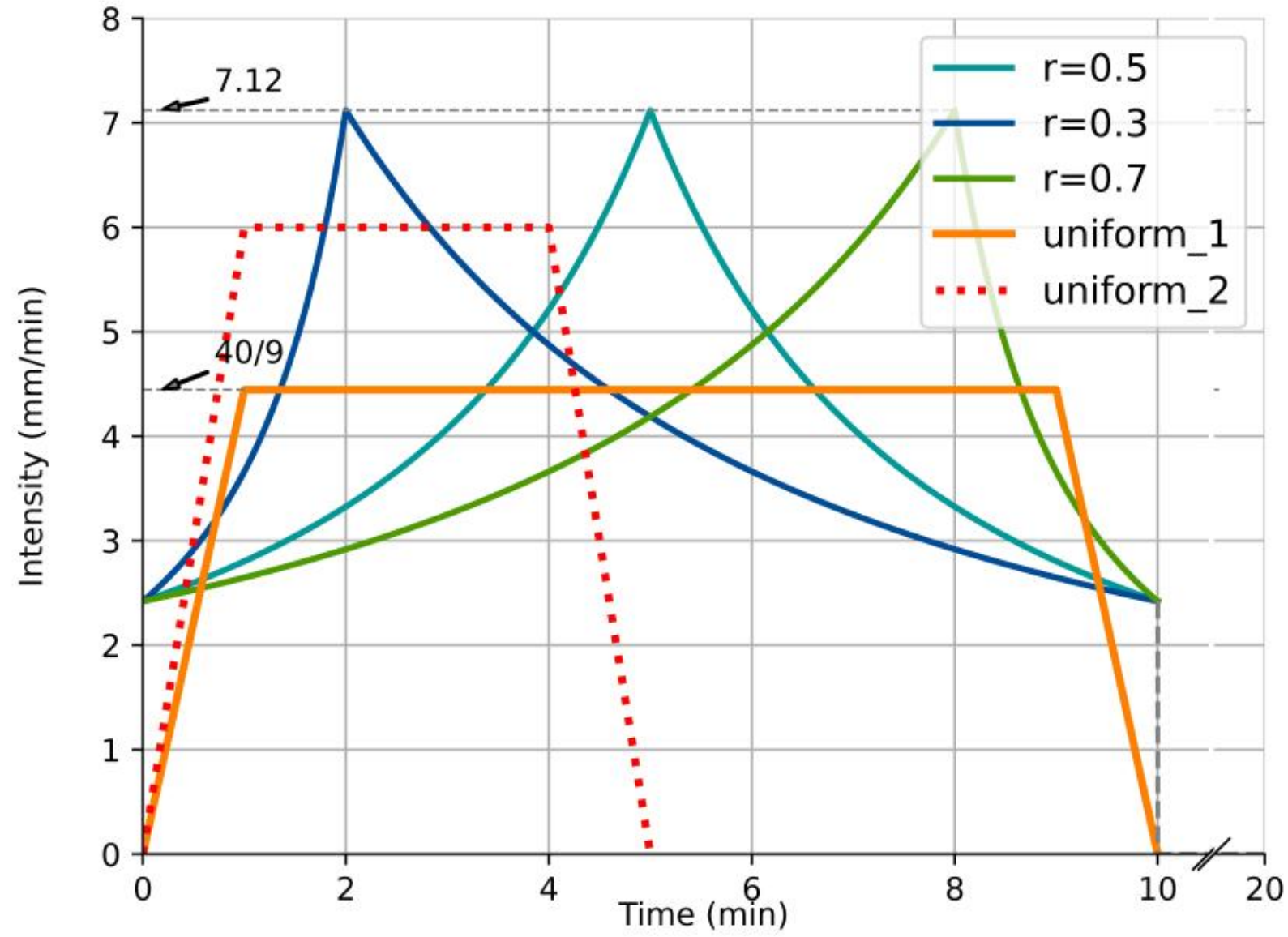

Fig.3. Four design rainfalls for Case 1 (solid lines) and one design rainfall for Case 2 (dotted line)

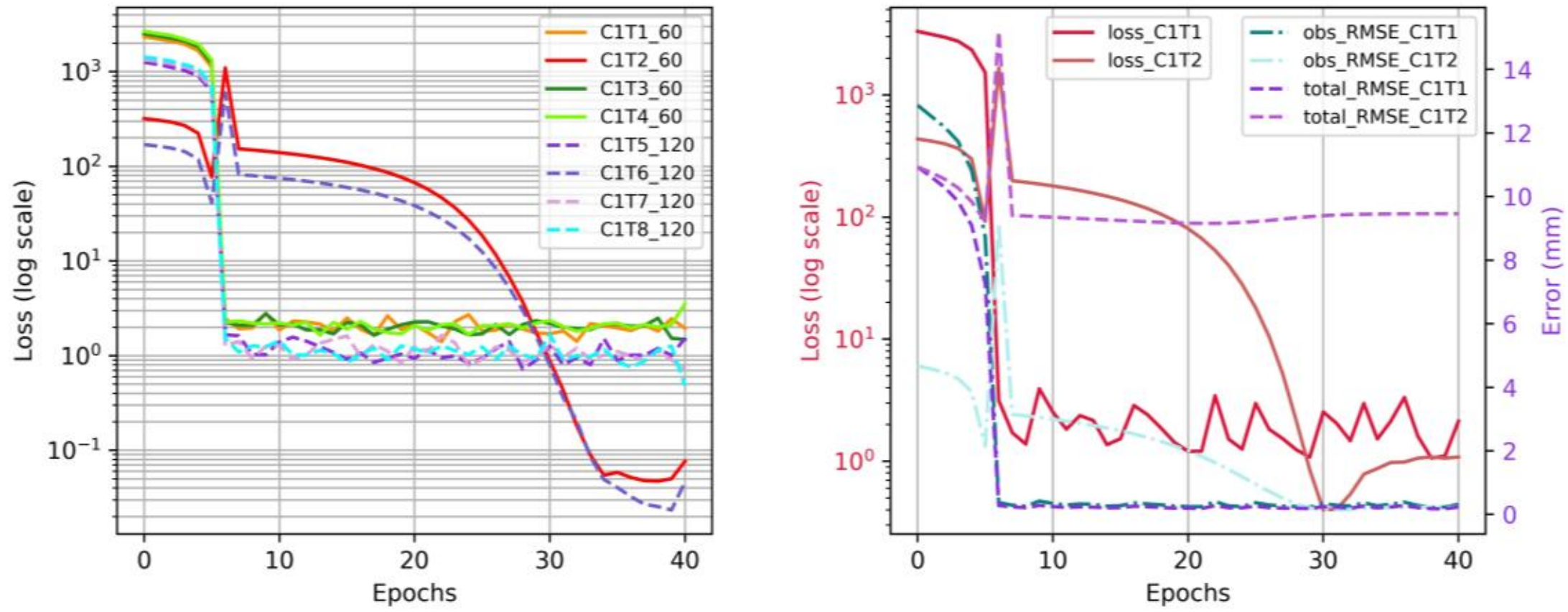

Fig.4. The loss functions of parameter calibration for 8 observation schemes under uniform rainfall (left); the test error of C1T1 and C1T2 observation schemes under the Chicago type rainfalls (right).

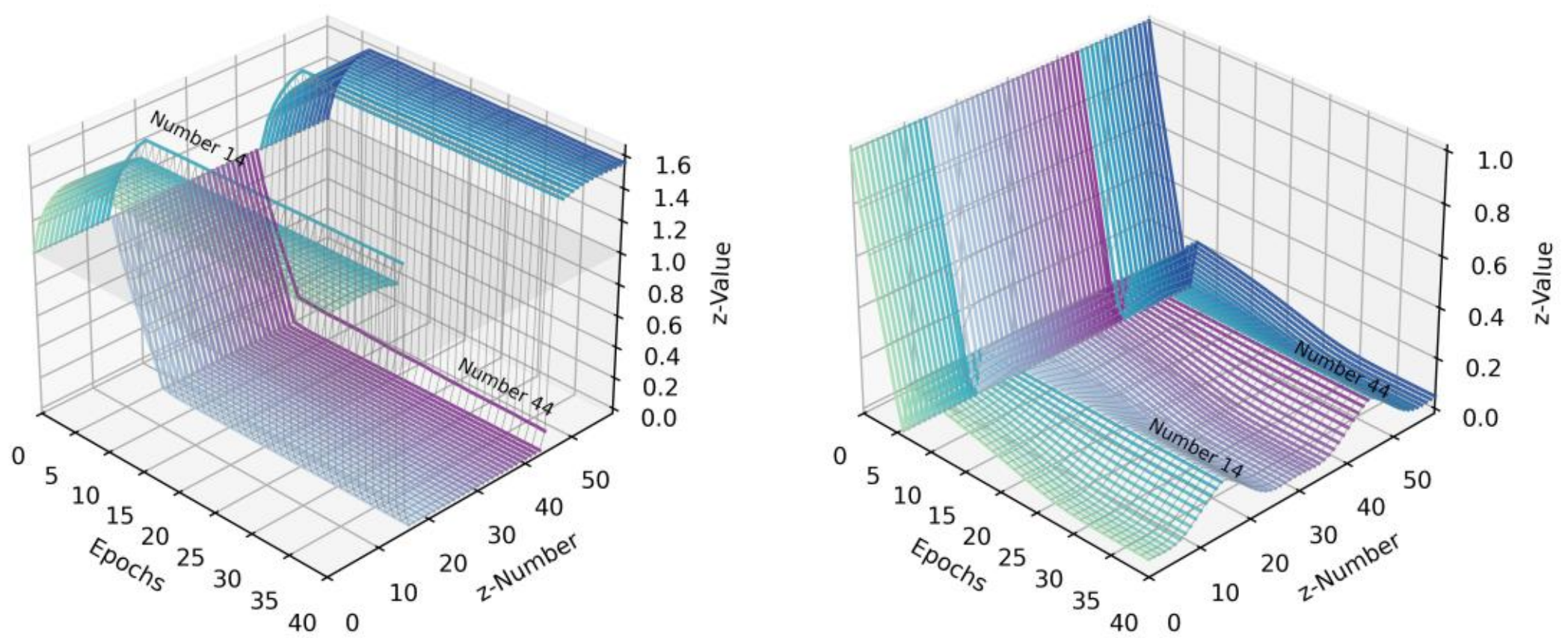

Fig.5. Posterior values of latent variables under the uniform rainfall for C1T1 scheme (left) and C1T2 scheme (right)

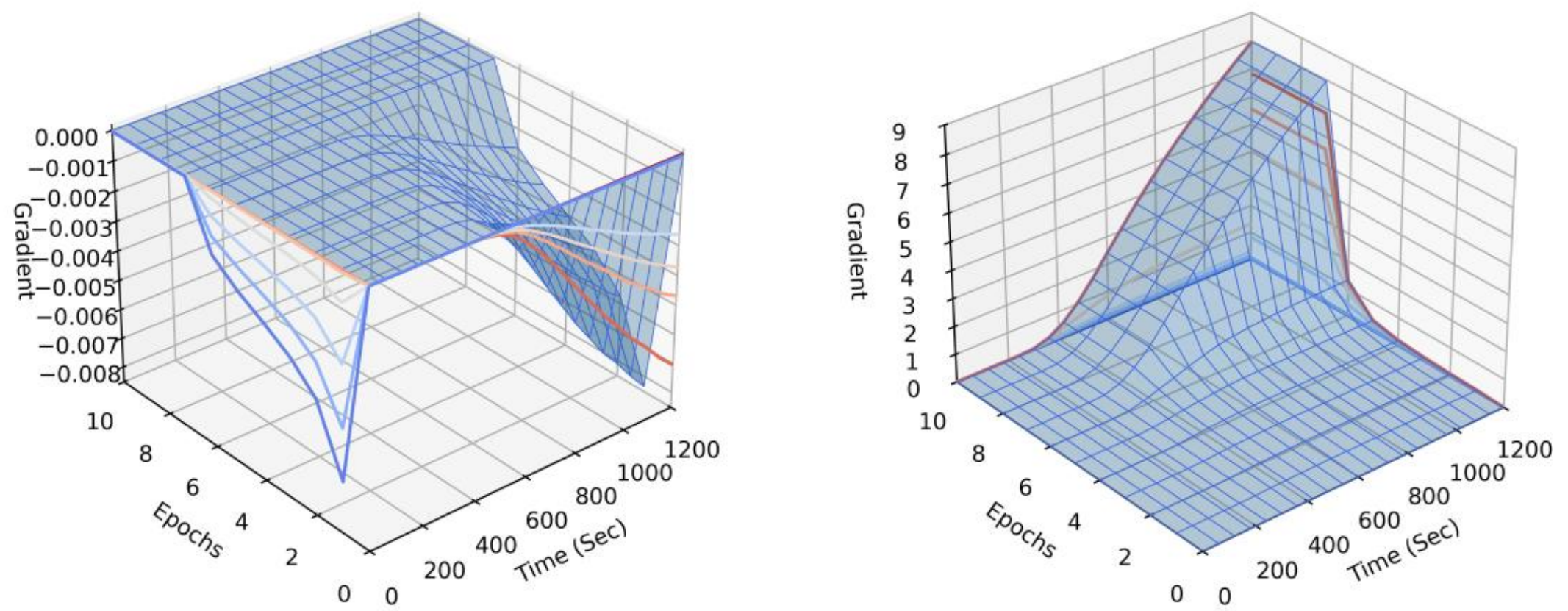

Fig.6. Gradient of the observation point 3 with respect to the latent variable 7 (left) and the latent variable 37 (right) under uniform rainfall in the C1T1 scheme

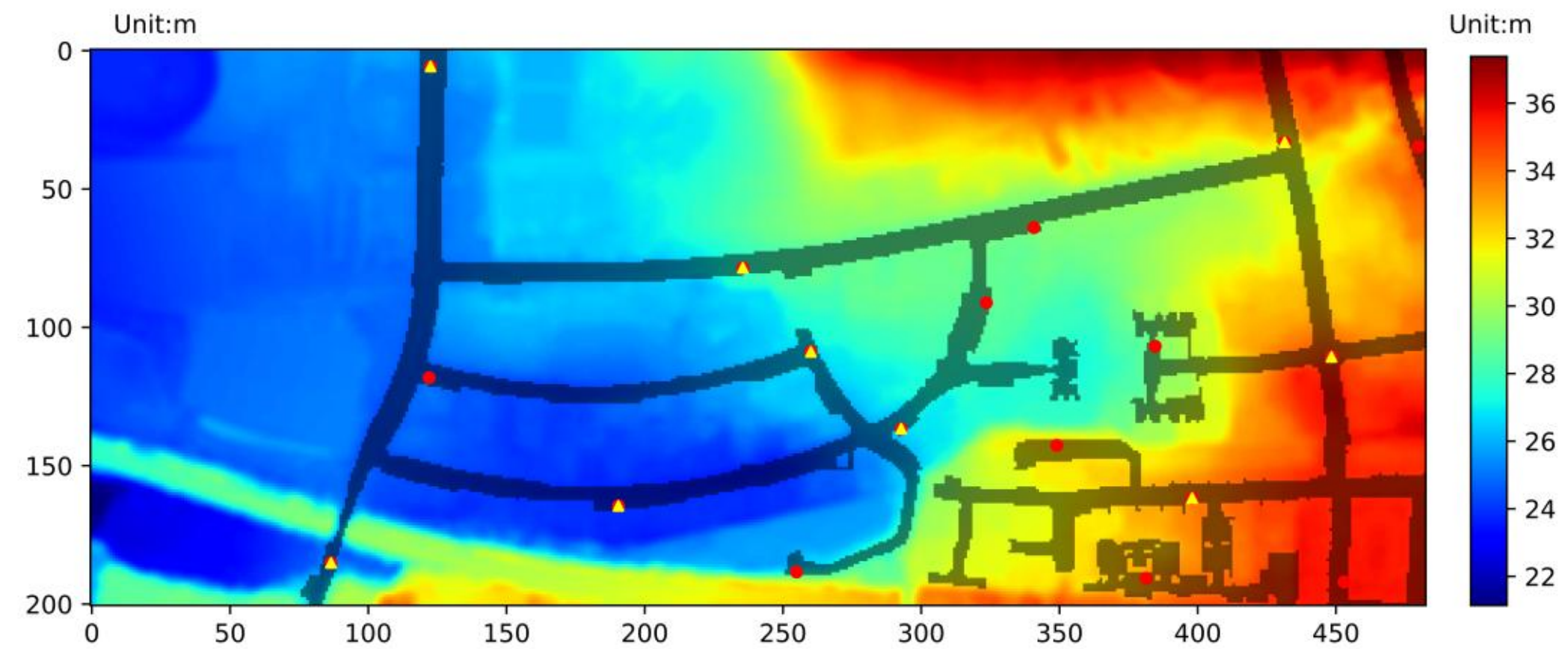

Fig.7. Digital Elevation Map of the Test 8A and two sets of observation schemes

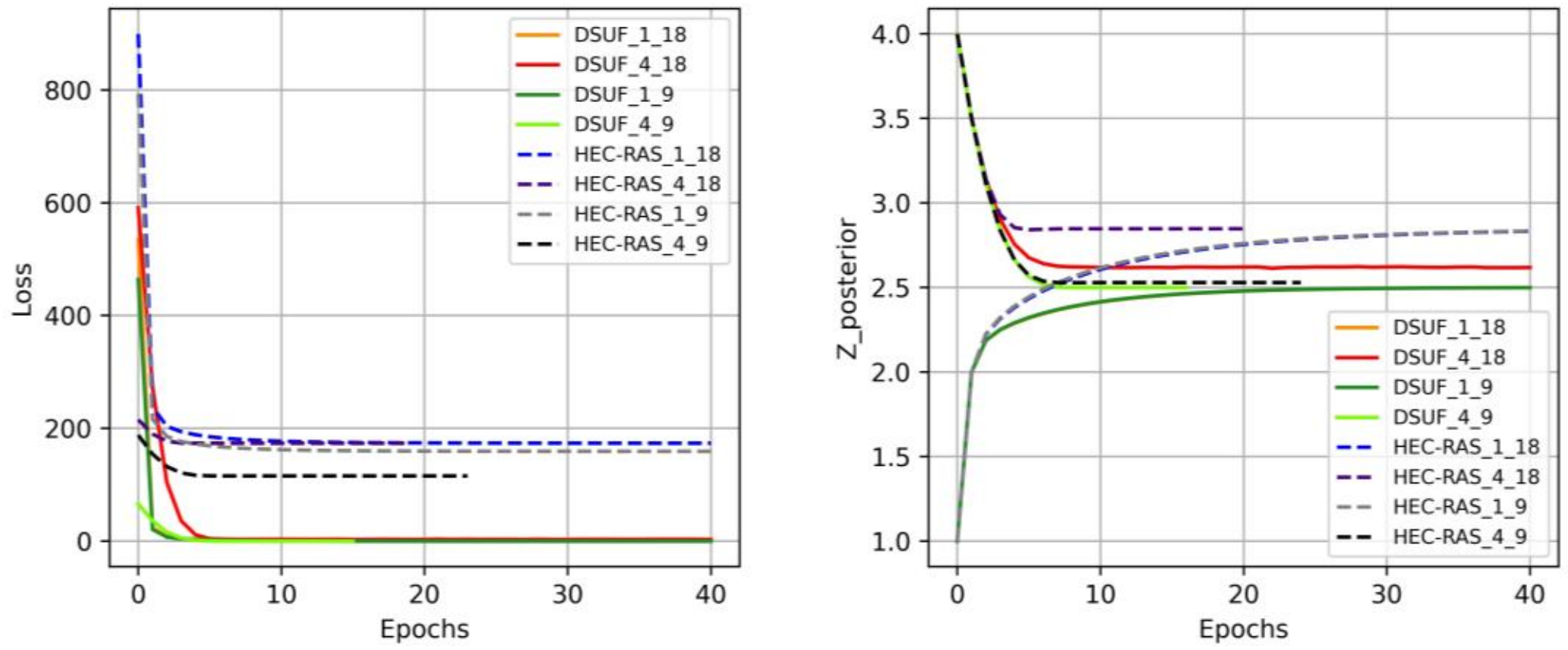

Fig.8. Loss functions of the 8 schemes (left) and posterior values of the latent variable (right) in Case 2.